\documentclass{article}

\usepackage[final]{neurips_2019}
\usepackage{float}

\usepackage[utf8]{inputenc}
\usepackage[T1]{fontenc}
\usepackage{hyperref}
\usepackage{url}
\usepackage{booktabs}
\usepackage{amsfonts}
\usepackage{nicefrac}
\usepackage{microtype}
\usepackage{graphicx}
\usepackage{xcolor}
\usepackage{lipsum}
\usepackage{subfig}
\usepackage{subfiles}
\usepackage{amssymb}
\usepackage{pifont}
\usepackage[most]{tcolorbox}
\newcommand{\cmark}{\ding{51}}%
\newcommand{\xmark}{\ding{55}}%
\usepackage{array}
\newcolumntype{L}{>{\arraybackslash}m{1.4cm}}
\newcolumntype{M}{>{\arraybackslash}m{1cm}}

\title{
  QAGAN: Adversarial Approach To Learning Domain Invariant Language Features
}

\author{
  Shubham Shrivastava \\
  Stanford University
  \thanks{email: \url{shubhams@stanford.edu}, the author is also affiliated with Ford Greenfield Labs. All correspondence should be addressed to \url{shubham@towardsautonomy.com}} \\
  \And
  Kaiyue Wang \\
  Stanford University
  \thanks{email: \url{garywang@stanford.edu}, the author is also affiliated with Waymo LLC}
}

\begin{document}

\bibliographystyle{unsrtnat}
\setcitestyle{square,numbers,comma}

\maketitle

\begin{abstract}
Training models that are robust to data domain shift has gained an increasing interest both in academia and industry \cite{awais2021adversarial}. Question-Answering language models, being one of the typical problem in Natural Language Processing (NLP) research, has received much success with the advent of large transformer models \cite{pearce2021comparative}. However, existing approaches mostly work under the assumption that data is drawn from same distribution during training and testing which is unrealistic and non-scalable in the wild.

In this paper, we explore adversarial training approach towards learning domain-invariant features so that language models can generalize well to \textit{out-of-domain} datasets. We also inspect various other ways to boost our model performance including data augmentation by paraphrasing sentences, conditioning \textit{end} of answer span prediction on the \textit{start} word, and carefully designed annealing function. Our initial results shows that in combination with these methods, we are able to achieve $15.2\%$ improvement in EM score and $5.6\%$ boost in F1 score on \textit{out-of-domain} validation dataset over the baseline. We also dissect our model outputs and visualize the model hidden-states by projecting them onto a lower-dimensional space, and discover that our specific adversarial training approach indeed encourages the model to learn domain invariant embedding and bring them closer in the multi-dimensional space. \footnote{Code implementation available here: \url{https://github.com/towardsautonomy/QAGAN}}
\end{abstract}

\section{Introduction}
Question answering in language modeling is a difficult task as it requires language models to understand the context and the question. Transformers based encoder-only models such as BERT \cite{devlin-etal-2019-bert} has had tremendous success in solving these tasks. These models are pre-trained on a tremendous amount of unlabeled data from BooksCorpus \cite{Zhu_2015_ICCV} and Wikipedia, and then fine-tuned on downstream tasks like question-answering. During fine-tuning, these models encode both question and context embedding together separated by a special separator token \texttt{[SEP]}, and then predicts for each context word probabilities for it being either a \textit{start} word or an \textit{end} word. Fine-tuning these models yield impressive results using only a single classification layer on top of pre-trained BERT.

While these models result in great performance on downstream tasks, they fail to generalize well across datasets. This is attributed by the inherent domain gap between various datasets which causes a large performance drop when tested on an unseen dataset. We attempt to minimize this domain gap and work towards building a model that is robust to domain shifts in dataset by learning domain-invariant features through adversarial training. We show through systematic evaluations that our approach actually helps the network close this gap and generalize better to the \texttt{out-of-domain} datasets.

We also condition our \textit{end\_logit} prediction on the \textit{start\_logit} through an \texttt{MLP} and a \texttt{Self-Attention} prediction head, and we find that in both cases it boosts the performance further. Our approach to minimizing the domain gap through adversarial training to confuse a discriminator working towards classifying the domain that each sample came from is intuitive and have proven to work in computer vision literature \cite{CycleGAN2017, Wei2018PersonTG, DBLP:conf/cvpr/JaipuriaZBACSMM20}. Additionally, we tried augmenting our data using paraphrasing by back translation and word substitution with close embedding   \cite{dataAugmentation2019}. We observed that such technique does help with generalization and over fitting issue and has positive improvement on out of domain classification task. 

Concretely, our main contributions can be summarized as:

\begin{itemize}
\setlength\itemsep{0.1em}
    \item We show through various qualitative and quantitative evaluations that adversarial training helps the model learn domain-invariant features across various datasets.
    \item We design an annealing function called \textit{heated \texttt{tanh} annealing}, and show that it improves the model performance.  
    \item We present a simple conditional \texttt{MLP} and \texttt{Self-Attention} prediction head for \textit{start} and \textit{end} logits prediction, and show through various experiments that they further helps the model perform better.
\end{itemize}


\section{Related Work}

\textbf{Domain Adversarial Training:} The idea of adversarial training was first proposed by \cite{goodfellowAdversarial} and is mainly used for task generation problem. Since then, learning domain invariant features through adversarial training framework has gain increasing attention in research community.

One method proposed by \cite{Ganin2016DomainAdversarialTO} trained 2 classifiers one to perform task specific classification task and the other, using the hidden state of the first classifier, predicts whether the data belong to source or target domain. Lee et al. \cite{lee-etal-2019-domain} implemented a multi-domain classifier for the question-answering task to correctly predict the domain each data sample came from, while the QA model objective was set to minimize the KL divergence between discriminator output and a uniform distribution.
Sato et al. \cite{sato-etal-2017-adversarial} expanded on the binary classification discriminator architecture with an additional neural layer for each domain, which captures domain-specific feature representations and applied it to dependency parsing tasks. 

\textbf{Data Augmentation:} Data augmentation has been shown to prevent model from over-fitting. Longpre et al. \cite{dataAugmentation2019} explored data augmentation for QA task by paraphrasing the question context through back translation. Along the same line, Garg et al. \cite{gargBert} used \texttt{BERT-MLM} to generate alternatives sentences for masked tokens. Wei et al. \cite{wei-zou-2019-eda} proposed some easy data augmentation techniques such as synonym replacement, random insert, random swap and deletion which are shown to improve performance of task with limited small data set. 

We take inspiration from these work and build a model that is able to learn domain-invariant features, and further boost its performance through several techniques such as conditional logits prediction, data augmentation, annealing, and fine-tuning.

\section{Approach}

We set out to explore the domain-gap within various QA datasets \cite{kwiatkowski-etal-2019-natural, trischler-etal-2017-newsqa,rajpurkar-etal-2016-squad,levy-etal-2017-zero,DuoRC,lai-etal-2017-race} by projecting higher-dimensional embedding extracted through a language model trained on humongous volume of unlabeled data onto a lower-dimensional space and then visually inspecting them. Specifically, we use a pre-trained DistilBERT \cite{sanh2020distilbert} model and project the hidden-states vectors onto a 2-dimensional space using t-SNE \cite{vanDerMaaten2008}. We observe that there indeed exists a domain-gap across these datasets as shown in figure \ref{fig:tsne_distilbert}. Furthermore, we see that when these networks are fine-tuned on a downstream task, it further segregates the embeddings farther apart to a point where data from each dataset is clustered in its own island of higher-dimensional space. This motivates our approach of adversarial training regime where a discriminator is trained with an objective of classifying the dataset domain of question-answer pair, and the language model works towards confusing the discriminator. We call our model \texttt{QAGAN}, and as shown in figure \ref{fig:tsne_qagan}, it does reduce the domain gap across datasets.

We work with $6$ datasets \cite{kwiatkowski-etal-2019-natural, trischler-etal-2017-newsqa,rajpurkar-etal-2016-squad,levy-etal-2017-zero,DuoRC,lai-etal-2017-race} in our work, for three of which (\texttt{indomain\_train}) we have $50k$ training samples per dataset, and for the other three (\texttt{oodomain\_train}) we have $127$ training samples as described in section \ref{sec:datasets}. Goal of our work is to train a model that generalizes well to \texttt{oodomain\_val} and \texttt{oodomain\_test} dataset.

\subsection{Baseline}
Our baseline system fine-tunes a pre-trained DistilBERT \cite{sanh2020distilbert} on all \texttt{indomain\_train} dataset. The loss function is set up as a combination of negative log likelihood for the start and end location.

\subsection{QAGAN}
For building our model, we take the huggingface's \cite{wolf-etal-2020-transformers} implementation of DistilBERT \cite{sanh2020distilbert} and build a simple MLP-based prediction head to attach on top of the pre-trained DistilBERT's sequence encoder output. We also take the discriminator model implementation from the work by Lee et al. \cite{lee-etal-2019-domain} and add support for training the model in adversarial fashion. The discriminator was implemented such that it works towards discriminating the dataset domain by operating on its \texttt{[CLS]} representation as it summarizes the whole context of the paragraph. We chose to use DistilBERT as our baseline to avoid longer GPU training time, less contribution towards global warming and to save those polar bears.

\begin{figure}[!h]
  \centering
  \subfloat[\texttt{qagan} with linear conditional prediction head]{\includegraphics[width=0.48\linewidth]{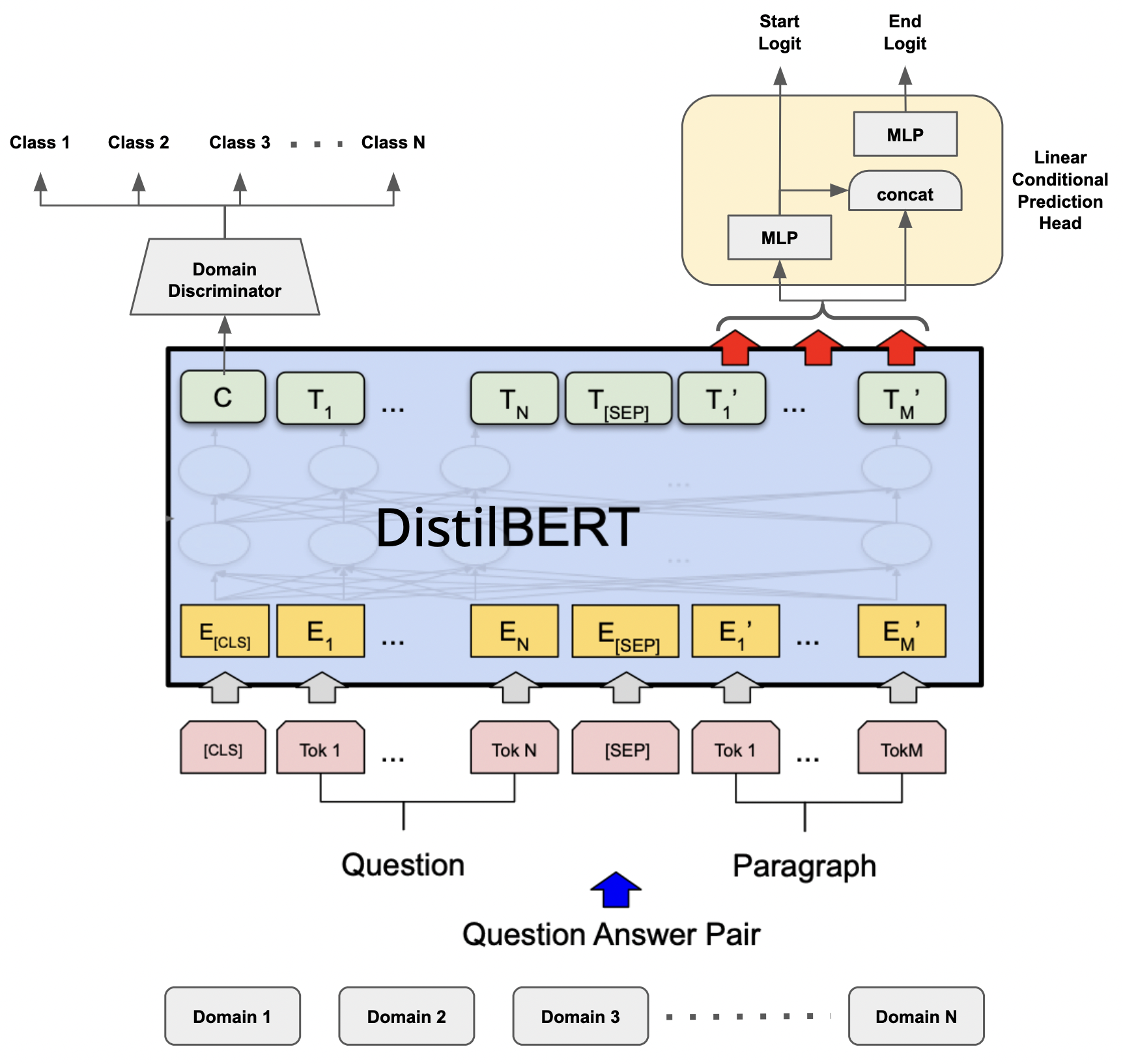}\label{fig:qagan-cond}}
  \hfill
  \subfloat[\texttt{qagan} with self-attention based conditional prediction head]{\includegraphics[width=0.48\linewidth]{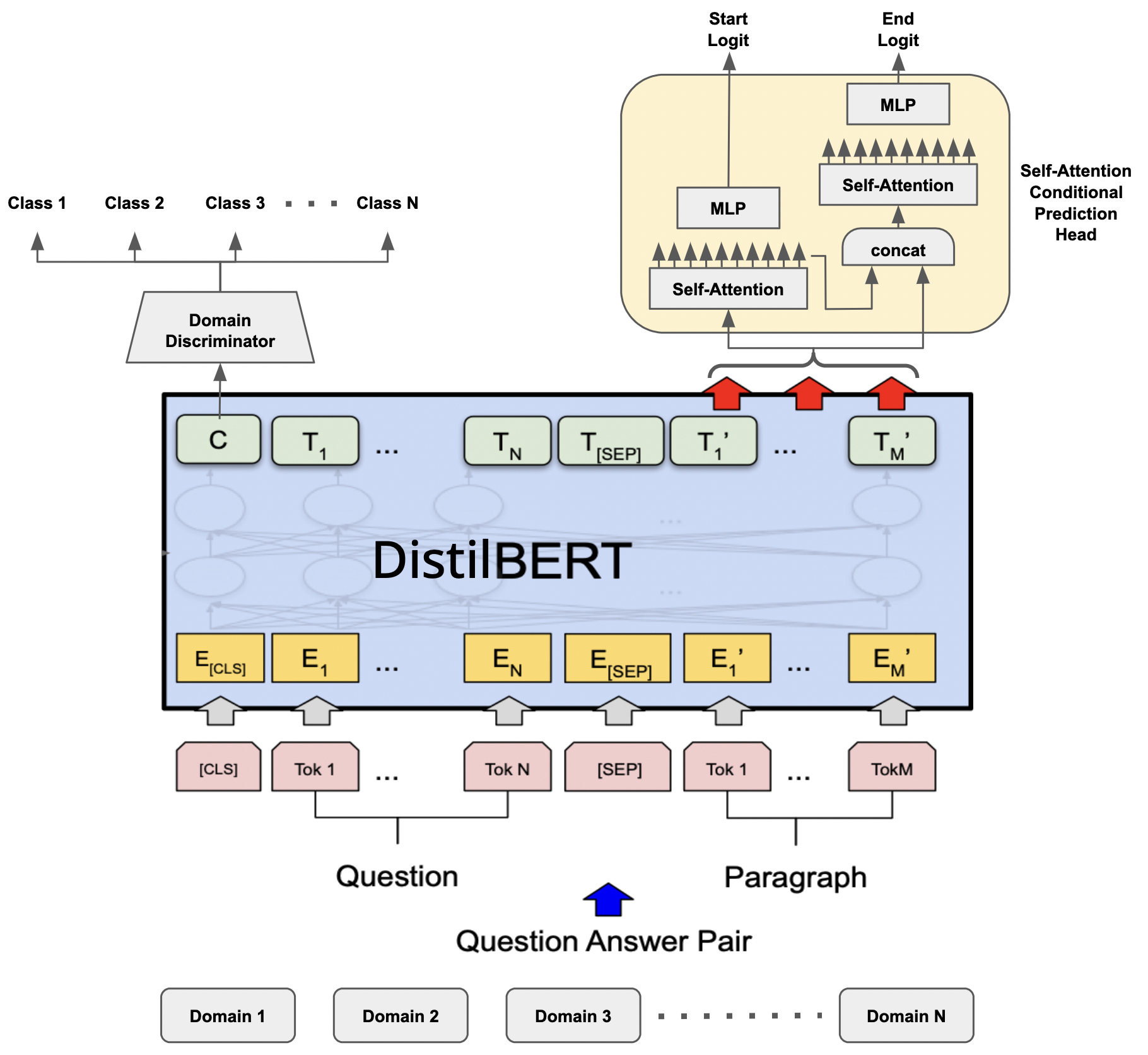}\label{fig:qagan-cond-att}}
  \caption{Variants of \texttt{QAGAN}}
  \label{fig:qagan}
\end{figure}

We also take inspiration from Wang et al. \cite{DBLP:journals/corr/WangJ16a} and design a conditional prediction heads to condition our \textit{end\_logits} prediction on \textit{start\_logits} and notice that it improves our model performance. Two prediction heads designed in our work were linear MLP and self-attention based, and while both improved our model performance, we find the linear conditional head to yield the best results. These two models are shown in figure \ref{fig:qagan}.

We train our \textit{question-answering} (QA) model by minimizing the \textit{negative-log-likelihood} loss between ground-truth and prediction. In addition, we train our discriminator to classify the dataset domain class correctly, while the QA model learns how to fool the discriminator into outputting random dataset class probability.

\vspace{-0.1in}
\begin{center}
\begin{equation}\label{eqn:qa_loss}
    L_{QA} = - \frac{1}{N} \sum^K_{k=1} \sum^{N_k}_{i = 1}[log P_{\theta} (\hat{y}_{i, s}^{(k)}| x_i^{(k)}, q_i^{(k)}) + log P_{\theta}(\hat{y}_{i,e}^{(k)}| x_i^{(k)}, q_i^{(k)})]
\end{equation}
\vspace{-0.1in}
\begin{equation}\label{eqn:adv_loss}
    L_{adv} = - \frac{1}{N} \sum^K_{k=1} \sum^{N_k}_{i = 1}{y}_{i, rand}^{(k)} log P_{\theta} (\hat{y}_{i, c}^{(k)}| h_{cls}^{(k)})
\end{equation}
\vspace{-0.1in}
\begin{equation}\label{eqn:disc_loss}
    L_{D} = - \lambda_2 \frac{1}{N} \sum^K_{k=1} \sum^{N_k}_{i = 1}y_{i, c}^{(k)}log P_{\theta} (\hat{y}_{i, c}^{(k)}| h_{cls}^{(k)})
\end{equation}
\end{center}

During QA model training we optimize the loss function, $L_{qagan} = L_{QA} + \lambda_1 L_{adv}$, and during the discriminator model training we optimize the loss function $L_{D}$, where $\lambda_1$ and $\lambda_2$ is a hyper-parameter chosen to control the importance of adversarial loss. Furthermore, $L_{adv}$ is optimized every $n_{fd}$ steps and $L_{D}$ is optimized every $n_{td}$ steps.

$N$, $\hat{y}_{i,s}$, and $\hat{y}_{i,e}$ in equation \ref{eqn:qa_loss} are
the total number of in-domain data, the start position and the end position of answer in the passage respectively.
$\hat{y}_{i,c}$ and $h_{cls}$ in equation \ref{eqn:adv_loss} are the class of the dataset and \texttt{[CLS]} hidden representations respectively. By adding this term in the loss function during training the QA model, it is encouraging the QA model to produce hidden states that would confuse the discriminator model such that it would output random probabilities for each class.

During our initial experiments we did not observe considerable performance improvement, and we hypothesize that at the beginning of training, discriminator does not have much information to go off from, and hence training it to correctly classify the domain while the QA model works towards confusing it ends up worsening the model performance. To counter this, we design an annealing function (\textit{heated tanh annealing}) described in section \ref{sec:anneal} which helps improve the network performance and also allows it to learn domain-invariant features. Additionally, we replaced the discriminator input with complete hidden states instead of just the \texttt{[CLS]} representation to examine its effects, and it did not seem to improve the performance.

Lee et al. \cite{lee-etal-2019-domain} used Kullback-Leiber divergence between uniform distribution and the distribution of class probability generated by the discriminator during QA model training which helped the model generate a representation such that it confuses the discriminator. In our work, we also replaced this loss function with \textit{negative log-likelihood} between the discriminator output and random classes, and it proved to improve the model performance slightly.

\subsubsection{Heated \texttt{Tanh} Annealing}\label{sec:anneal}

\vspace{-0.1in}
\begin{figure}[!h]
  \centering
  \subfloat[$n_{ws}$=$10k$, $n_{max}$=$250k$]{\includegraphics[width=0.4\linewidth]{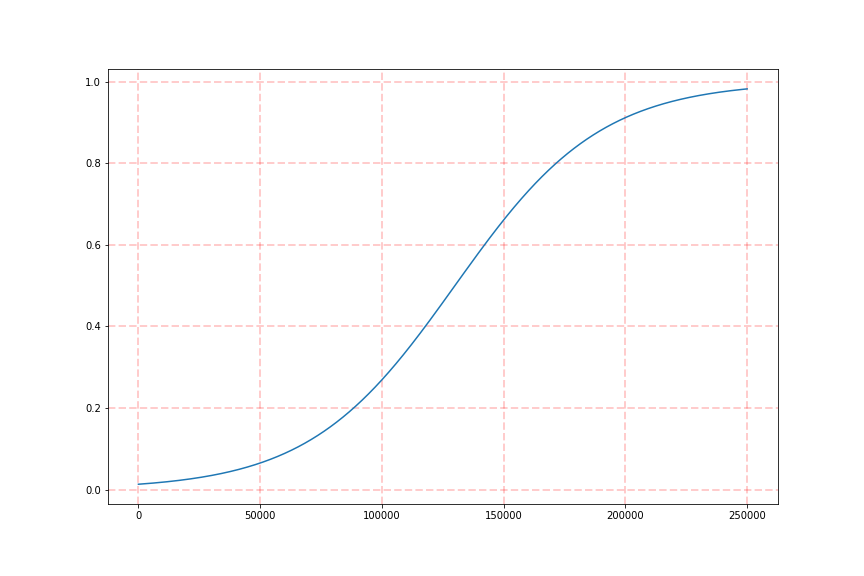}\label{fig:heated_tanh_annealing_10k}}
  \subfloat[$n_{ws}$=$150k$, $n_{max}$=$250k$]{\includegraphics[width=0.4\linewidth]{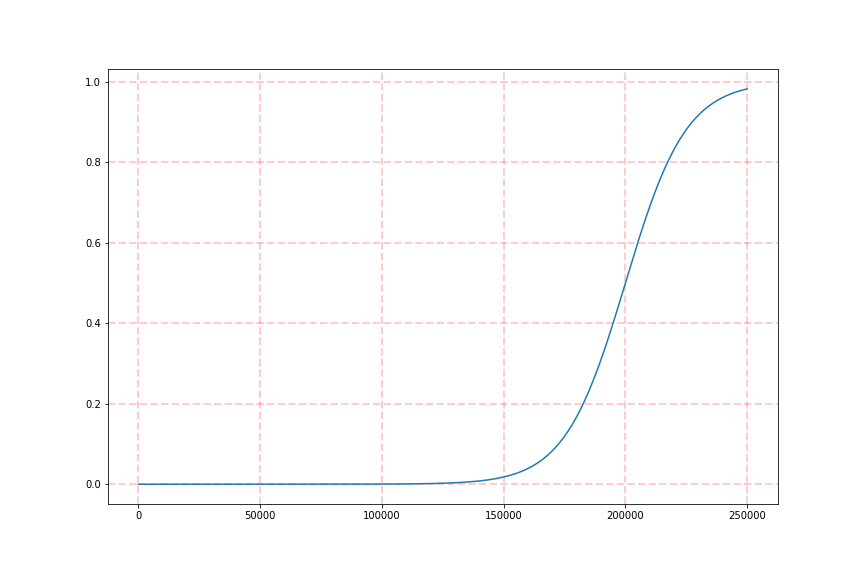}\label{fig:heated_tanh_annealing_150k}}
  \caption{Heated \texttt{Tanh} Annealing for optimizing \textit{discriminator}.}
  \label{fig:heated_tanh_annealing}
\end{figure}

We designed an annealing function as given by equation \ref{eqn:heated_tanh_anneal} and illustrated in figure \ref{fig:heated_tanh_annealing}. This function allows the discriminator to initially learn a good model for domain classification, and slowly after $n_{ws}$ warm-up steps, the model starts using the discriminator output as a proxy for learning domain-invariant features. We find that weighting the \textit{fake discriminator loss} with this function resulted in an improved performance as expected.

\begin{equation}\label{eqn:heated_tanh_anneal}
    f_{anneal}(z) = \frac{tanh(2*\frac{2z-n_{max}-n_{ws}}{n_{max}-n_{ws}}) + 1}{2}
\end{equation}

Where, $n_{max}$ is maximum number of steps, and $n_{ws}$ is the number of warm-up steps.

\subsubsection{Conditional QA Prediction Head}
We designed two conditional heads for \textit{start} and \textit{end} logits prediction. These heads are illustrated in figures \ref{fig:qagan-cond} and \ref{fig:qagan-cond-att} respectively. 

\textbf{Linear Conditional Head:} Sequence output of DistilBERT is passed through an MLP to produce \texttt{start\_logit} which is then concatenated with the same sequence and passed through another MLP to produce \texttt{end\_logit}.  

\textbf{Self-Attention Conditional Head:} Sequence output of DistilBERT is passed through a \textit{self-attention} layer followed by an MLP to produce \texttt{start\_logit}. Output of this \textit{self-attention} layer is concatenated with the sequence output and fed through another \textit{self-attention} layer followed by another MLP layer to then produce \texttt{end\_logit}.  

\subsubsection{Data Augmentation}
For data augmentation, we tried 2 different techniques to generate more data in order to regularize the model for more robustness.

\textbf{Back Translation:} \
We used a trained machine translation model MarianMTModel \cite{junczys-dowmunt-etal-2018-marian} provided by hugging face to translate the context and question into German and translate it back to English as a paraphrasing mechanism. In order to maintain the quality of the back translation, we slice the context around the answer text with an added padding. We also make sure that the sliced index does not happen within a sentence by heuristically looking for sentence ending punctuation. Moreover, to maintain the translated sentence quality, we calculate a perplexity score using pre-trained GPT-2 model \cite{Radford2019LanguageMA} and only take the translated text that doesn't deviate from original text. 

It is also observed that after translation, the answer text will get morphed and can no longer be found in the translated text. We just skip the generated text under these circumstances.

The paraphrasing reduces the dependency on domain specific sentence structure. For example, we see that the sentence:
    
\fbox{a tiny girl the size of the old woman's thumb.}

gets paraphrased to:

\fbox{a little girl the size of the thumb of the old woman.}

\textbf{Word Replacement:}
To increase the robustness of word choices, we also does word replacement where we randomly replace words with its counter parts that is close in embedding space.  

\begin{figure}[H]
  \centering
    \subfloat[Validation plot on model with data augmentation (red) and model without data augmenation (blue)]{\includegraphics[width=0.8\linewidth]{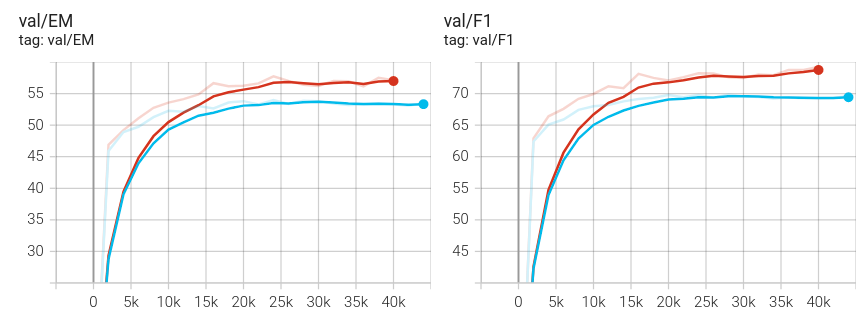}}\label{fig:data_augmented_training}
\end{figure}

\subsection{Fine-tuning:} 
After the models have been trained on \texttt{in-domain}-train set, we also fine-tune them on the \texttt{out-of-domain}-train set, and that brings about $3.8\%$ boost in EM score and $3.3\%$ boost in F1 score while tested on \texttt{out-of-domain}-validation set. This however, has worsens the performance when the model is initially trained on both (\texttt{in-domain}-train + \texttt{out-of-domain}-train) datasets, and rightfully so, because the model will overfit to the small \texttt{out-of-domain}-train split during fine-tuning.

\section{Experimental Details}

We built various variants of \texttt{QAGAN} and performed experiments to test our hypothesis. 

\begin{itemize}
\setlength\itemsep{0.1em}
    \item Starting from the discriminator loss function during QA model training, we experimented with minimizing \textit{KL-divergence} between discriminator classification and a uniform distribution, and \textit{negative-log-likelihood} between discriminator classification output and random dataset domain class (\texttt{$D_{obj}$} column in table \ref{tab:qagan-results}).
    \item We designed several prediction heads including a simple MLP (\textit{$P_{head}={mlp}$}), conditional MLP (\textit{$P_{head}={cmlp}$}), and a conditional self-attention (\textit{$P_{head}={csat}$}) based sub-network.
    \item Another option during our model training is to use either \texttt{[CLS]} representation as an input to the discriminator, or use the complete \texttt{hidden-states} representation, and in our experiments we found the former to yield better results.  
    \item Using insights from our domain-gap analysis, we will also experiment with enforcing on better generalization by constraining hidden state to conform to normal distribution (h\_{kld} column in table \ref{tab:qagan-results}).
    \item We also designed an annealing function to aid the training of our discriminator, and experimented with and without this function to evaluate its effectiveness (\textit{anneal} column in table \ref{tab:qagan-results}).
    \item We trained our model by combining the augmented data to evaluate the effectiveness of data augmentation(\textit{aug} column in table \ref{tab:qagan-results})
    \item For some of our experiments we also included the \textit{out-of-domain train} set into the complete training set to analyze its effect on final model performance (\textit{ood\_train} column in table \ref{tab:qagan-results})
\end{itemize}

For all our experiments, we used a learning rate of $3e-5$, and trained our model for a total of $3$ epochs. Hyper-parameters used in our loss functions were: $\lambda_1=0.5$, $\lambda_2=0.5$, $n_{ws}=1000$, $n_{max}=250k$,  $n_{td}=1$,  $n_{fd}=2$. Training was done on a machine with single NVIDIA RTX 3090 GPU, and takes about $53$ minutes per epoch. 

\subsection{Dataset Description}\label{sec:datasets}

To setup our experiment framework, we take a subset of the MRQA \cite{fisch2019mrqa} \textit{in-domain} and \textit{out-of-domain} dataset, statistics for which is shown in table \ref{tab:mrqa-subset-statistics} \footnote{This dataset split can be downloaded from \href{https://drive.google.com/file/d/1Fv2d30hY-2niU7t61ktnMsi_HUXS6-Qx/view?usp=sharing}{here}}. This includes 3 \textit{in-domain} reading comprehension datasets (Natural Questions \cite{kwiatkowski-etal-2019-natural}, NewsQA \cite{trischler-etal-2017-newsqa} and SQuAD \cite{rajpurkar-etal-2016-squad}), each have a sample size of 50K for training the QA system. It also includes 3 small \textit{out-of-domain} datasets (RelationExtraction \cite{levy-etal-2017-zero}, DuoRC \cite{DuoRC}, RACE \cite{lai-etal-2017-race}) with size of 127 for fine-tuning. 
To preprocess the data, we converted each (question, paragraph) into multiple chunks of size 384 with a stride of 128 in order for it to fit the DistilBERT model input dimension. 

\begin{table}[!h]
    \centering
    \caption{Statistics for datasets used for building the QA system (MRQA-subset). \textbf{Question Source} and \textbf{Passage Source} refer to data sources from which the questions and passages were obtained.}
    \begin{tabular}{llllll}
    \toprule
    \multicolumn{1}{c}{\textbf{Dataset}} & \multicolumn{1}{c}{\textbf{Question Source}} & \multicolumn{1}{c}{\textbf{Passage Source}} & \multicolumn{1}{c}{\textbf{Train}} & \multicolumn{1}{c}{\textbf{Validation}} & \multicolumn{1}{c}{\textbf{Test}}  \\ 
    \midrule
    \multicolumn{6}{c}{\texttt{in-domain datasets}} \\
    \midrule 
    SQuAD\cite{rajpurkar-etal-2016-squad} & Crowdsourced & Wikipedia & 50,000 & 10,507 & - \\
    NewsQA\cite{trischler-etal-2017-newsqa} & Crowdsourced & News articles & 50,000 & 4,212 & - \\
    Natural Questions\cite{kwiatkowski-etal-2019-natural} & Search logs & Wikipedia & 50,000 & 12,836 & - \\
    SQuAD Aug & Crowdsourced & Wikipedia & 20,000 & - & - \\
    NewsQA Aug & Crowdsourced & News articles & 20,000 & - & - \\
    Natural Questions Aug & Search logs & Wikipedia & 20,000 & - & - \\
    \midrule 
    \multicolumn{6}{c}{\texttt{oo-domain datasets}} \\
    \midrule 
    DuoRC\cite{DuoRC} & Crowdsourced & Movie reviews & 127 & 126 & 1,248 \\
    RACE\cite{lai-etal-2017-race} & Teachers & Examinations & 127 & 126 & 419 \\
    RelationExtraction\cite{levy-etal-2017-zero} & Synthetic & Wikipedia & 127 & 126 & 2,693 \\
    DuoRC Aug & Crowdsourced & Movie reviews & 314 & - & - \\
    RACE Aug & Teachers & Examinations & 240 & - & - \\
    RelationExtraction Aug & Synthetic & Wikipedia & 250 & - & - \\

    \bottomrule
    
    \end{tabular}
    \label{tab:mrqa-subset-statistics}
\end{table}

As introduced in previous section, we also explored various data augmentation techniques including paraphrasing by back-translation introduced in \cite{DBLP:journals/corr/abs-1912-02145} and semantic-preserving perturbation techniques introduced in \cite{ribeiro-etal-2018-semantically} which generates an additional 20K for each indomain datasets and 1K for each oodomain datasets.

For the sake of completion, we also used the full MRQA dataset \cite{fisch2019mrqa} for training our model and report the result on \textit{out-of-domain} dataset. Distribution for this full dataset is shown in table.

\begin{table}[!h]
    \centering
    \caption{Statistics for full MRQA \cite{fisch2019mrqa} dataset used for training and validation. \textbf{Question Source} and \textbf{Passage Source} refer to data sources from which the questions and passages were obtained.}
    \begin{tabular}{lllll}
    \toprule
    \multicolumn{1}{c}{\textbf{Dataset}} & \multicolumn{1}{c}{\textbf{Question Source}} & \multicolumn{1}{c}{\textbf{Passage Source}} & \multicolumn{1}{c}{\textbf{Train}} & \multicolumn{1}{c}{\textbf{Validation}} \\ 
    \midrule
    \multicolumn{5}{c}{\texttt{in-domain datasets}} \\
    \midrule 
    SQuAD\cite{rajpurkar-etal-2016-squad} & Crowdsourced & Wikipedia & 86,588 & 10,507 \\
    NewsQA\cite{trischler-etal-2017-newsqa} & Crowdsourced & News articles & 74,160 & 4,212 \\
    TriviaQA\cite{joshi-etal-2017-triviaqa} & Crowdsourced & Trivia enthusiasts & 61,688 & 7,785 \\
    SearchQA\cite{https://doi.org/10.48550/arxiv.1704.05179} & Search logs & Web crawling & 117,384 & 16,980 \\
    HotpotQA\cite{yang-etal-2018-hotpotqa} & Crowdsourced & Wikipedia & 72,928 & 5,904 \\
    Natural Questions\cite{kwiatkowski-etal-2019-natural} & Search logs & Wikipedia & 104,071 & 12,836 \\
    \midrule 
    \multicolumn{5}{c}{\texttt{oo-domain datasets}} \\
    \midrule 
    BioASQ\cite{TSA+12} & Crowdsourced & Bio-medical literature & - & 1,504 \\
    DROP\cite{dua-etal-2019-drop} & Crowdsourced & Wikipedia + NFL & - & 1,503 \\
    DuoRC\cite{DuoRC} & Crowdsourced & Movie reviews & - & 1,501 \\
    RACE\cite{lai-etal-2017-race} & Teachers & Examinations & - & 674 \\
    RelationExtraction\cite{levy-etal-2017-zero} & Synthetic & Wikipedia & - & 2,948 \\
    TextbookQA\cite{8100054} & Crowdsourced & Middle school curricula & - & 1,503 \\
    \bottomrule
    
    \end{tabular}
    \label{tab:mrqa-statistics}
\end{table}

\section{Qualitative Analysis}
\begin{figure}[!h]
  \centering
 \subfloat[\texttt{distilbert-features}]{\includegraphics[width=0.33\linewidth]{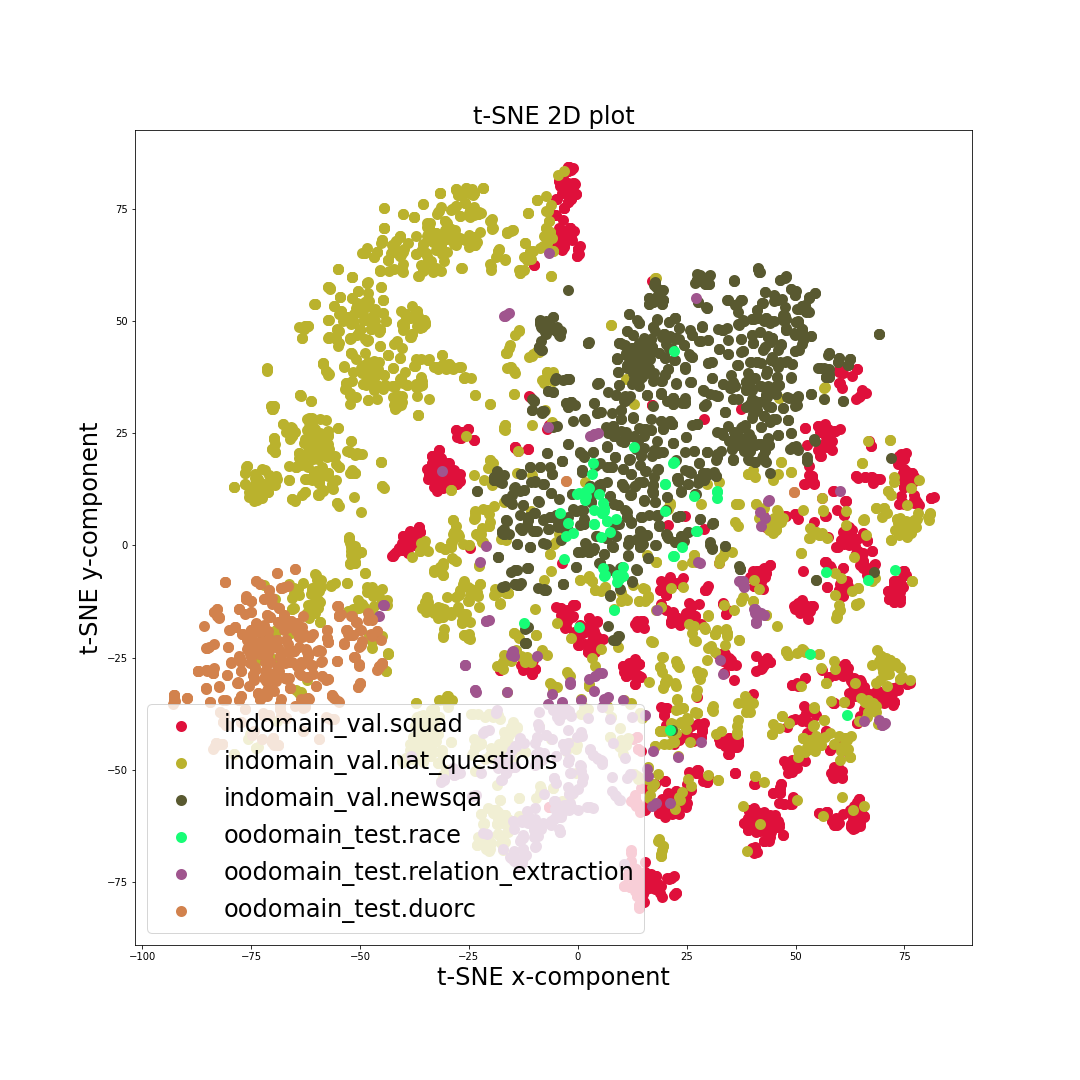}\label{fig:tsne_distilbert}}
  \subfloat[\texttt{baseline}]{\includegraphics[width=0.33\linewidth]{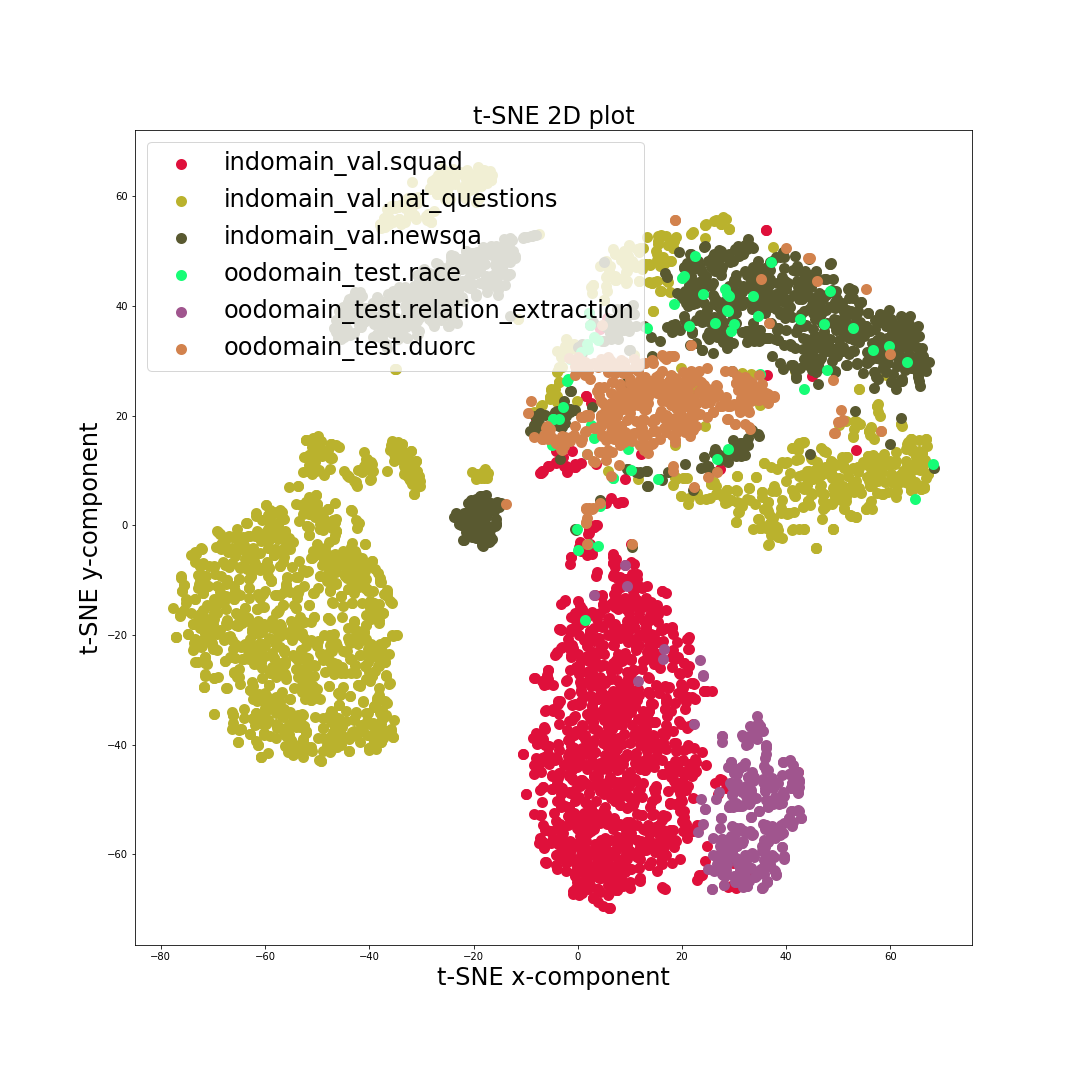}\label{fig:tsne_baseline}}
  \subfloat[\texttt{qagan}]{\includegraphics[width=0.33\linewidth]{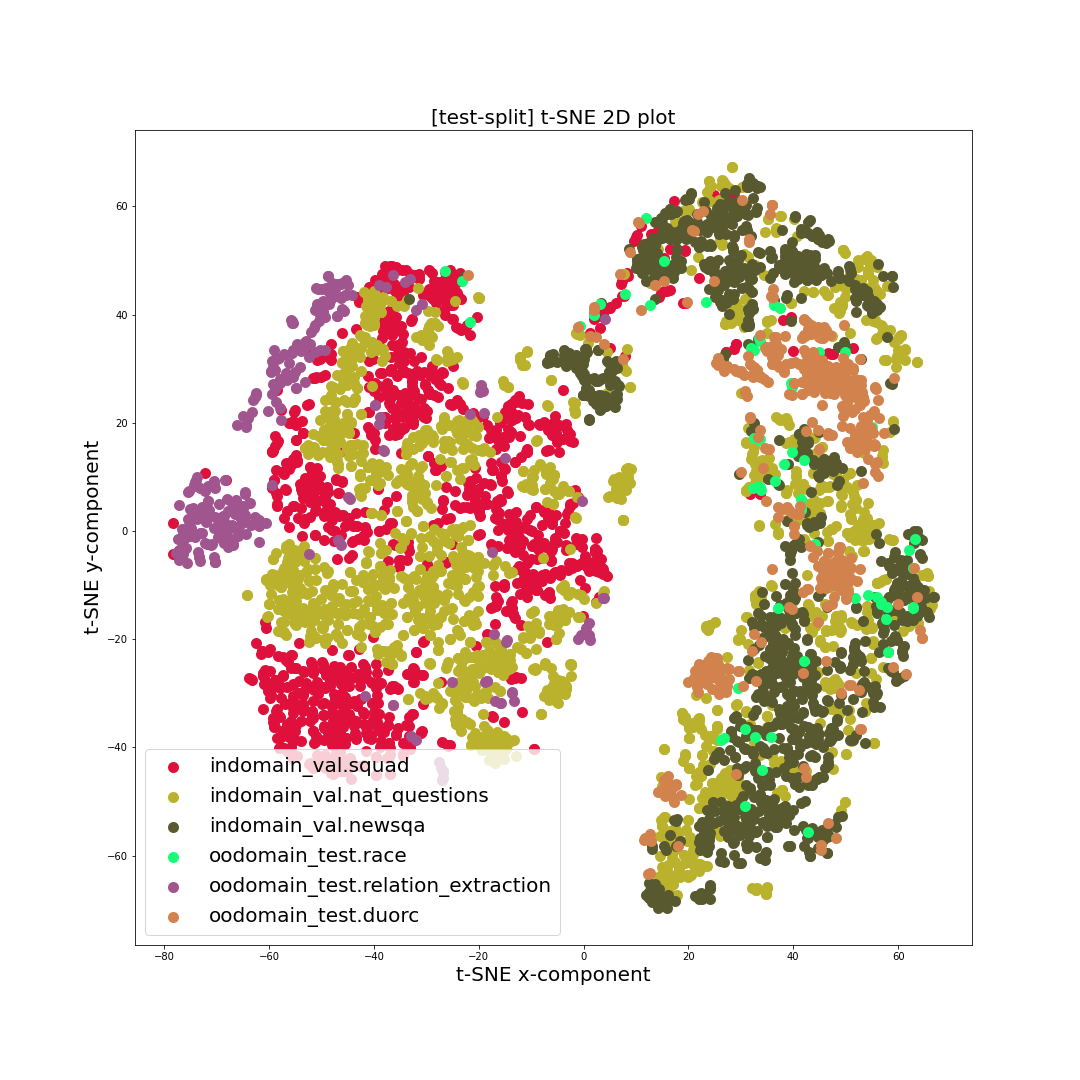}\label{fig:tsne_qagan}}
  \caption{[MRQA-subset] Domain-gap across various QA dataset embeddings and the effectiveness of applying our method towards learning domain-invariant features.}
\end{figure}

\begin{figure}[!h]
  \centering
 \subfloat[\texttt{distilbert-features}]{\includegraphics[width=0.33\linewidth]{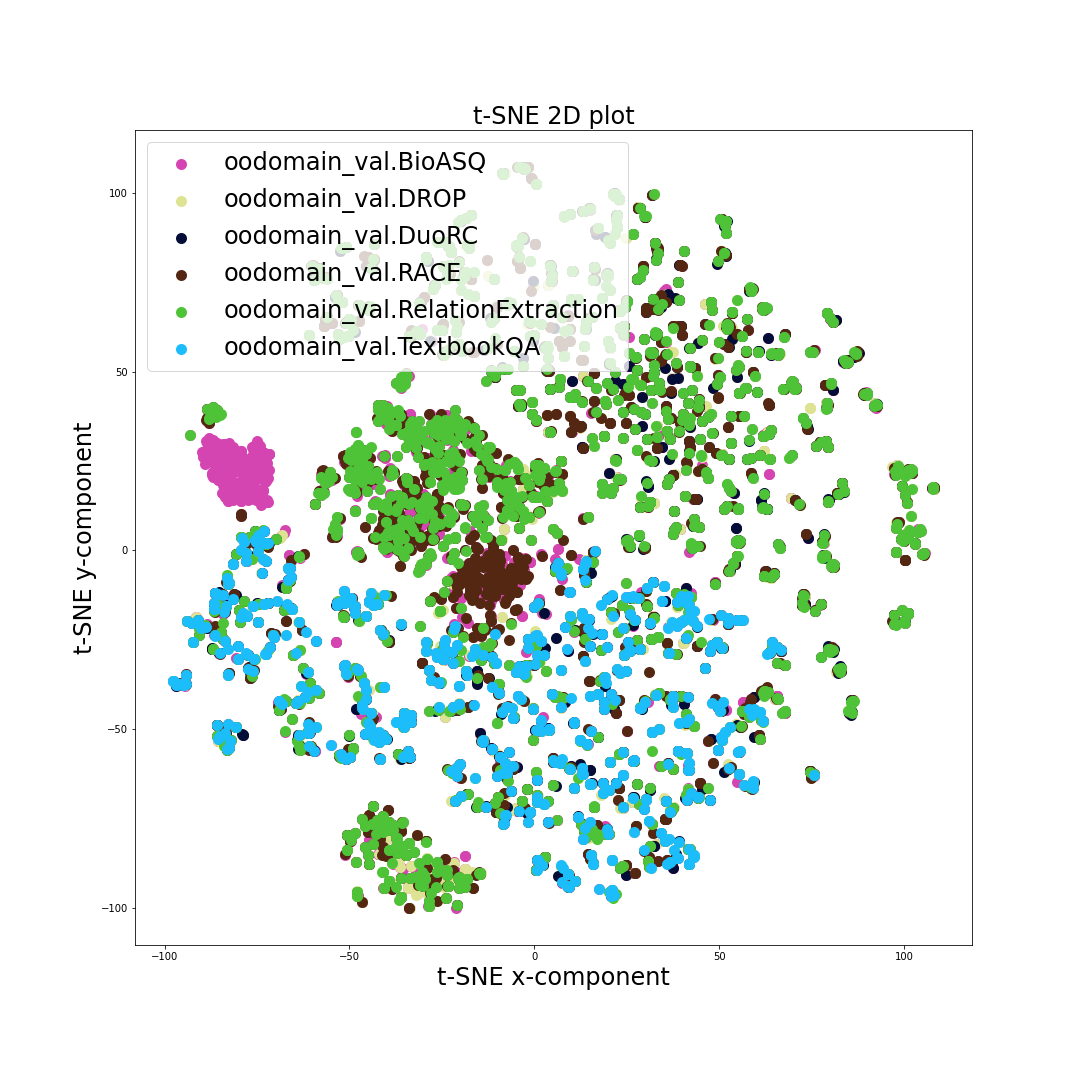}\label{fig:tsne_distilbert_mrqa}}
  \subfloat[\texttt{baseline}]{\includegraphics[width=0.33\linewidth]{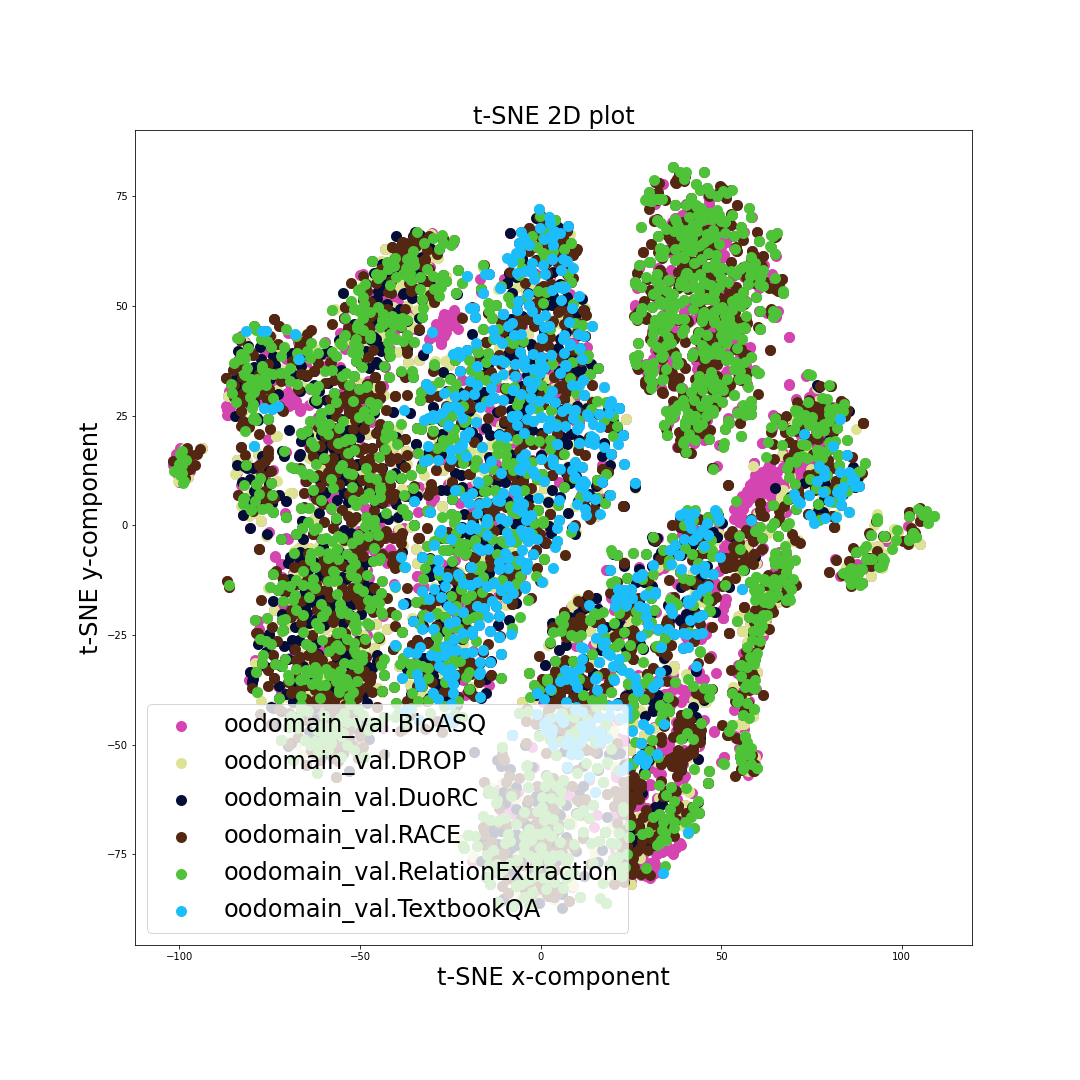}\label{fig:tsne_baseline_mrqa}}
  \subfloat[\texttt{qagan}]{\includegraphics[width=0.33\linewidth]{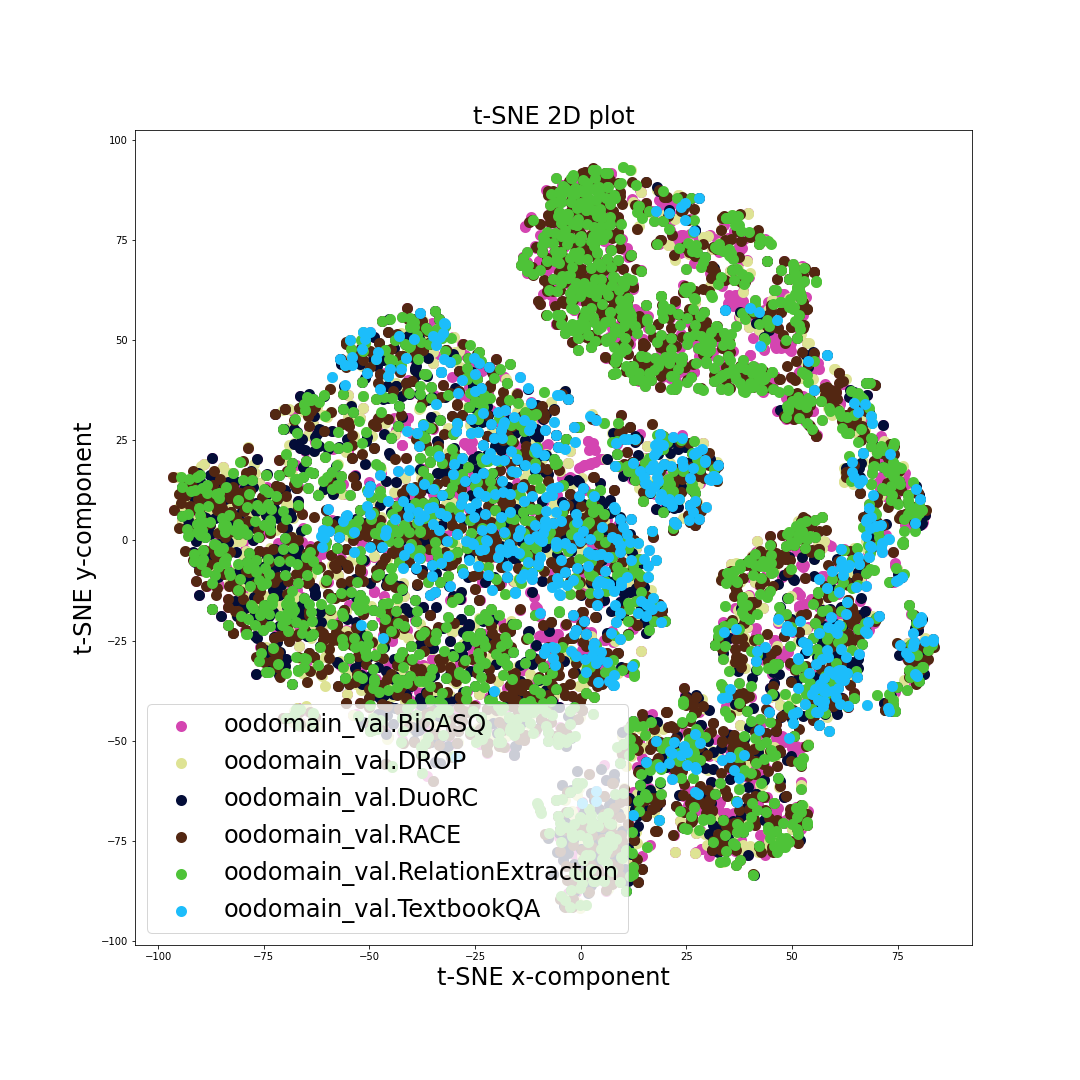}\label{fig:tsne_qagan_mrqa}}
  \caption{[Full MRQA] Domain-gap across various QA dataset embeddings and the effectiveness of applying our method towards learning domain-invariant features.}
\end{figure}

\textbf{Effectiveness of Adversarial Training:} Figure \ref{fig:tsne_distilbert} illustrates the domain-gap across various datasets, and is extracted through a pre-trained \textit{DistilBERT} \cite{sanh2020distilbert} model. We train a \textit{baseline} QA model on \textit{indomain\_train} dataset which is formulated by adding an \texttt{MLP} prediction head on top of the \textit{DistilBERT} model and analyze the embeddings. As seen in the plot (figure \ref{fig:tsne_baseline}, embeddings for each dataset coming from this trained model is clustered individually and clearly shows that the domain-gap is accentuated even more with this setting. Lastly, with our approach, \texttt{QAGAN}, we find that although multi-modal, various dataset embeddings coexists in similar high-dimensional space as shown in figure \ref{fig:tsne_qagan}. This solidifies our hypothesis that the adversarial training setting indeed helps the model learn domain-invariant features. Figures \ref{fig:tsne_distilbert}, \ref{fig:tsne_baseline}, and \ref{fig:tsne_qagan} shows embeddings for \texttt{MRQA-subset} datasets. A similar plot has been shown in figures \ref{fig:tsne_distilbert_mrqa}, \ref{fig:tsne_baseline_mrqa}, and \ref{fig:tsne_qagan_mrqa} for the full \texttt{MRQA} validation datasets, and again we observe the same pattern.

\textbf{Quality of Predicted Answers:}
We show a few random qualitative samples predicted by our model and further analyze the quality of predictions in appendix (section \ref{sec:appendix}). In particular, we observe that the model does extremely well on \textit{named entities}, but suffers badly with contexts requiring some common reasoning.

\subsection{Quantitative Results}

\textbf{Evaluation method:} We evaluated our method via two metrics: \textbf{Exact Match (EM)} score and \textbf{F1} score. Exact Match (EM) is a strict binary measure that tells us whether the predicted span of words exactly match the reference. F1 measure, on the other hand, is a more relaxed metric and is the harmonic mean of precision and recall. Both metrics are widely used in the evaluation of question and answering task. \cite{rajpurkar-etal-2016-squad}

For \texttt{MRQA-Subset} dataset split (table \ref{tab:mrqa-subset-statistics}), we evaluate \textbf{F1} and \textbf{EM} scores on all validation and test sets and report in tables \ref{tab:qagan-results} and \ref{tab:qagan-results-test} respectively, whereas, for the full \texttt{MRQA} dataset (table \ref{tab:mrqa-statistics}), we evaluate these scores on each validation dataset in the validation split.

\begin{table}[!h]
    \centering
    \caption{Experimental results of our method on \texttt{MRQA-subset} \textit{validation} datasets. Blue color highlights best performance without fine-tuning on \texttt{$oodomain\_train$} dataset or including \texttt{$oodomain\_train$} in the \textit{train} set. $P_{head} \rightarrow \texttt{prediction head}$, $h\_kld \rightarrow \texttt{KLD loss on hidden-states}$, $D_{input} \rightarrow \texttt{discriminator input}$, $D_{obj} \rightarrow \texttt{discriminator objective}$, $ood\_train \rightarrow \texttt{oodomain\_train set included in training}$, $aug \rightarrow \texttt{data augmentation}$. Prediction Heads: $\texttt{mlp} \rightarrow \texttt{linear MLP head}$; $\texttt{cmlp} \rightarrow \texttt{condition MLP head}$; $\texttt{csat} \rightarrow \texttt{conditional self-attention head}$.}
    \begin{tabular}{l|l|l|l|l|l|l|l|l|cc|cc}
    \toprule
    \multicolumn{1}{c|}{\textbf{Method}} & \multicolumn{1}{c|}{$P_{head}$} & \multicolumn{1}{c|}{\rotatebox{90}{finetune}} & \rotatebox{90}{anneal} & \rotatebox{90}{h\_kld} & \rotatebox{90}{ood\_train} & \rotatebox{90}{aug} & $D_{input}$ & $D_{obj}$ & \multicolumn{2}{c|}{\texttt{ind\_val}} & \multicolumn{2}{c}{\texttt{}{ood\_val}} \\ 
    \cline{10-13} 
    \multicolumn{1}{c|}{} & & & & & & & & & \textbf{F1} & \textbf{EM} & \textbf{F1} & \textbf{EM} \\
    \midrule 
    \texttt{baseline} & \texttt{mlp} & \xmark & - & - & \xmark & \xmark & - & - & \textcolor{blue}{70.49} & \textcolor{blue}{54.48} & 48.29 & 30.89 \\
    \texttt{baseline} & \texttt{mlp} & \cmark & - & - & \xmark & \xmark & - & - & - & - & 49.68 & 34.03 \\
    \midrule 
    \texttt{qagan} & \texttt{mlp} & \xmark & \cmark & \xmark & \xmark & \xmark & [CLS] & KLD  & 70.10 & 54.24 & 46.56 & 31.15 \\
    \texttt{qagan} & \texttt{mlp} & \cmark & \cmark & \xmark & \xmark & \xmark & [CLS] & KLD  & - & - & 47.38 & 33.25 \\
    \texttt{qagan} & \texttt{mlp} & \xmark & \cmark & \xmark & \xmark & \xmark & [hidn] & NLL  & 68.88 & 52.69 & 46.95 & 30.89 \\
    \texttt{qagan} & \texttt{mlp} & \cmark & \cmark & \xmark & \xmark & \xmark & [hidn] & NLL  & - & - & 48.46 & 34.03 \\
    \texttt{qagan} & \texttt{mlp} & \xmark & \cmark & \xmark & \xmark & \xmark & [CLS] & NLL  & 69.85 & 53.84 & 46.92 & 31.68 \\
    \texttt{qagan} & \texttt{mlp} & \cmark & \cmark & \xmark & \xmark & \xmark & [CLS] & NLL  & - & - & 49.16 & 34.03 \\
    \texttt{qagan} & \texttt{csat} & \xmark & \cmark & \xmark & \xmark & \xmark & [CLS] & NLL  & 69.79 & 53.67 & 47.32 & 31.15 \\
    \texttt{qagan} & \texttt{csat} & \cmark & \cmark & \xmark & \xmark & \xmark & [CLS] & NLL  & - & - & 48.87 & 34.55\\
    \texttt{qagan} & \texttt{cmlp} & \xmark & \xmark & \xmark & \xmark & \xmark & [CLS] & NLL  & 70.01 & 54.06 & 49.30 & 32.98 \\
    \texttt{qagan} & \texttt{cmlp} & \xmark & \cmark & \cmark & \xmark & \xmark & [CLS] & NLL  & 69.85 & 54.09 & 47.88 & 30.89 \\
    \texttt{qagan} & \texttt{cmlp} & \cmark & \cmark & \cmark & \xmark & \xmark & [CLS] & NLL  & - & - & 49.05 & 32.20 \\
    \texttt{qagan} & \texttt{cmlp} & \xmark & \cmark & \xmark & \xmark & \xmark & [CLS] & NLL  & 70.00 & 53.84 & \textcolor{blue}{49.38} & \textcolor{blue}{34.29} \\
    \texttt{qagan} & \texttt{cmlp} & \xmark & \cmark & \xmark & \cmark & \xmark & [CLS] & NLL  & 69.56 & 53.80 & 50.25 & 35.08 \\
    \texttt{qagan} & \texttt{cmlp} & \xmark & \cmark & \xmark & \cmark & \cmark & [CLS] & NLL  & 73.21 & 55.13 & 50.49 & \textbf{35.90} \\
    \texttt{qagan} & \texttt{cmlp} & \cmark & \cmark & \xmark & \xmark & \xmark & [CLS] & NLL  & - & - & \textbf{51.00} & 35.60 \\
    \bottomrule
    \end{tabular}
    \label{tab:qagan-results}
\end{table}

\begin{table}[!h]
    \centering
    \caption{Experimental results of our method on \texttt{MRQA-subset} \textit{test} datasets. $P_{head} \rightarrow \texttt{prediction head}$, $D_{obj} \rightarrow \texttt{discriminator objective}$, $ood\_train \rightarrow \texttt{oodomain\_train set included in training}$, $aug \rightarrow \texttt{data augmentation}$. Prediction Head: $\texttt{cmlp} \rightarrow \texttt{condition MLP head}$. \textit{heated \texttt{tanh} annealing} was used for all experiments.}
    \begin{tabular}{l|l|l|l|l|l|l|cc}
    \toprule
    \multicolumn{1}{c|}{\textbf{Method}} & \multicolumn{1}{c|}{$P_{head}$} & \rotatebox{90}{finetune} & \rotatebox{90}{ood\_train} & \rotatebox{90}{aug} & $D_{input}$ & $D_{obj}$ & \multicolumn{2}{c}{\texttt{oodomain\_test}} \\ 
    \cline{8-9} 
    \multicolumn{1}{c|}{} & & & & & & & \textbf{F1} & \textbf{EM} \\
    \midrule 
    \texttt{qagan} & \texttt{cmlp} & \cmark & \xmark & \xmark & [CLS] & NLL  & 58.193 & 40.665 \\
    \texttt{qagan} & \texttt{cmlp} & \xmark & \cmark & \xmark & [CLS] & NLL  & 58.671 & 41.009 \\
    \texttt{qagan} & \texttt{cmlp} & \xmark & \cmark & \cmark & [CLS] & NLL  & \textbf{58.898} & \textbf{42.362} \\
    \bottomrule
    \end{tabular}
    \label{tab:qagan-results-test}
\end{table}

Adversarial training helps our model learn domain-invariant features as evident by figure \ref{fig:tsne_qagan}. These domain-invariant features help the model generalize well to \textit{out-of-domain} dataset and hence perform better on the validation set. We train our network with various configurations of parameters and notice a consistent improvement in F1 and EM score of \textit{out-of-domain} validation dataset (ood\_val) as we keep adding components introduced in our work including \textit{conditional prediction head}, \textit{heated \texttt{tanh} annealing}, \textit{negative-log-likelihood} training of the discriminator during QA model training, and data augmentation. While we do not notice any improvement in F1 and EM score for \textit{in-domain} validation dataset, the scores for our best model on \textit{out-of-domain} validation set after finetuning our model on \textit{out-of-domain} train set is pretty significant ($5.6\%$ improvement in F1 score and $15.2\%$ improvement in EM score).

We also report evaluation results for the full \texttt{MRQA} dataset and show that we our method outperforms the baseline on \textit{out-of-domain} validation datasets on average for both \textit{F1} and \textit{EM} score.

\begin{table}[!h]
    \centering
    \caption{Experimental results of our method on \texttt{MRQA} \textit{validation} datasets. $P_{head} \rightarrow \texttt{prediction head}$, $D_{input} \rightarrow \texttt{discriminator input}$, $D_{obj} \rightarrow \texttt{discriminator objective}$. Prediction Heads: $\texttt{mlp} \rightarrow \texttt{linear MLP head}$; $\texttt{cmlp} \rightarrow \texttt{condition MLP head}$; $\texttt{csat} \rightarrow \texttt{conditional self-attention head}$.}
    \begin{tabular}{l|l|l|p{0.9em}p{0.9em}|p{0.9em}p{0.9em}|p{0.9em}p{0.9em}|p{0.9em}p{0.9em}|p{0.9em}p{0.9em}|p{0.9em}p{0.9em}|p{0.9em}p{0.9em}}
    \toprule
    \multicolumn{1}{c|}{\rotatebox{50}{\textbf{Method}}} & \multicolumn{1}{c|}{\rotatebox{90}{$P_{head}$}} & \multicolumn{1}{c|}{\rotatebox{90}{$D_{input}$}} & \multicolumn{2}{c|}{\rotatebox{90}{\texttt{BioASQ}\cite{TSA+12}}} & \multicolumn{2}{c|}{\rotatebox{90}{\texttt{DROP}\cite{dua-etal-2019-drop}}} & \multicolumn{2}{c|}{\rotatebox{90}{\texttt{DuoRC}\cite{DuoRC}}} & \multicolumn{2}{c|}{\rotatebox{90}{\texttt{RACE}\cite{lai-etal-2017-race}}} & \multicolumn{2}{c|}{\rotatebox{90}{\texttt{RelationExtraction}\cite{levy-etal-2017-zero}}} & \multicolumn{2}{c}{\rotatebox{90}{\texttt{TextbookQA}\cite{8100054}}} & \multicolumn{2}{c}{\rotatebox{90}{Average}} \\ 
    \cline{4-17}

    \multicolumn{1}{c|}{} & & & \textbf{F1} & \textbf{EM} & \textbf{F1} & \textbf{EM} & \textbf{F1} & \textbf{EM} & \textbf{F1} & \textbf{EM} & \textbf{F1} & \textbf{EM} & \textbf{F1} & \textbf{EM} & \textbf{F1} & \textbf{EM} \\
    \midrule 
    \rotatebox{50}{\texttt{baseline}} & \rotatebox{90}{\texttt{mlp}} & - & \rotatebox{90}{50.01} & \rotatebox{90}{34.40} & \rotatebox{90}{39.04} & \rotatebox{90}{27.29} & \rotatebox{90}{42.98} & \rotatebox{90}{31.72} & \rotatebox{90}{49.48} & \rotatebox{90}{33.73} & \rotatebox{90}{51.41} & \rotatebox{90}{35.56} & \rotatebox{90}{41.86} & \rotatebox{90}{30.21} & \rotatebox{90}{45.79} & \rotatebox{90}{32.15} \\
    
    \midrule 
    \rotatebox{50}{\texttt{qagan}} & \rotatebox{90}{\texttt{mlp}} & \rotatebox{90}{[CLS]} & \rotatebox{90}{49.61} & \rotatebox{90}{33.74} & \rotatebox{90}{37.40} & \rotatebox{90}{25.56} & \rotatebox{90}{42.05} & \rotatebox{90}{30.19} & \rotatebox{90}{49.10} & \rotatebox{90}{33.19} & \rotatebox{90}{50.65} & \rotatebox{90}{34.58} & \rotatebox{90}{40.12} & \rotatebox{90}{28.68} & \rotatebox{90}{44.82} & \rotatebox{90}{30.99} \\
    
    \midrule 
    \rotatebox{50}{\texttt{qagan}} & \rotatebox{90}{\texttt{mlp}} & \rotatebox{90}{[hidn]} & \rotatebox{90}{48.16} & \rotatebox{90}{33.43} & \rotatebox{90}{36.81} & \rotatebox{90}{25.94} & \rotatebox{90}{41.93} & \rotatebox{90}{31.42} & \rotatebox{90}{48.19} & \rotatebox{90}{33.24} & \rotatebox{90}{49.87} & \rotatebox{90}{34.62} & \rotatebox{90}{40.40} & \rotatebox{90}{29.94} & \rotatebox{90}{44.22} & \rotatebox{90}{31.43} \\
    
    \midrule 
    \rotatebox{50}{\texttt{qagan}} & \rotatebox{90}{\texttt{csat}} & \rotatebox{90}{[CLS]} & \rotatebox{90}{49.46} & \rotatebox{90}{33.70} & \rotatebox{90}{38.44} & \rotatebox{90}{26.56} & \rotatebox{90}{42.48} & \rotatebox{90}{30.79} & \rotatebox{90}{48.95} & \rotatebox{90}{33.25} & \rotatebox{90}{50.56} & \rotatebox{90}{34.53} & \rotatebox{90}{42.08} & \rotatebox{90}{30.07} & \rotatebox{90}{45.32} & \rotatebox{90}{31.48} \\
    
    \midrule 
    \rotatebox{50}{\texttt{qagan}} & \rotatebox{90}{\texttt{cmlp}} & \rotatebox{90}{[CLS]} & \rotatebox{90}{50.17} & \rotatebox{90}{34.41} & \rotatebox{90}{38.86} & \rotatebox{90}{26.67} & \rotatebox{90}{43.31} & \rotatebox{90}{31.92} & \rotatebox{90}{49.80} & \rotatebox{90}{34.11} & \rotatebox{90}{51.11} & \rotatebox{90}{35.04} & \rotatebox{90}{42.68} & \rotatebox{90}{31.27} & \rotatebox{90}{\textbf{45.99}} & \rotatebox{90}{\textbf{32.23}} \\
    
    \bottomrule
    \end{tabular}
    \label{tab:qagan-mrqa-results}
\end{table}

\section{Conclusion}
We presented a method for training a question-answering language model in an adversarial fashion and showed through various experiments that it helps the model generalize well to \textit{out-of-domain} dataset. We supplemented the discriminator model training with an annealing function, carefully designed to first let it learn correct classification task and then slowly help the language model learn domain-invariant features. Furthermore, we condition \textit{end} logits prediction on \textit{start} logits which facilitates a great deal of model improvement. Through systematic experiments with various configurations, we find that our model, \texttt{QAGAN}, with a conditional linear prediction head provides us the best results and achieves $5.6\%$ improvements in F1 score and $15.2\%$ improvement in EM score over the baseline. We analyze the higher-dimensional embeddings produced by our model and confirm that our method of adversarial training indeed helps the language model learn domain-invariant features and generalizes well towards \textit{out-of-domain} datasets.

\bibliography{references}

\clearpage
\pagenumbering{Alph}
\setcounter{section}{0}
\renewcommand{\thesection}{A\arabic{section}}  
\setcounter{figure}{0}
\renewcommand{\thefigure}{S\alph{figure}}
\setcounter{table}{0}
\renewcommand{\thetable}{S\alph{table}}
\setcounter{equation}{0}
\renewcommand{\theequation}{S\alph{equation}}
\onecolumn
\begin{center}
{ \LARGE \textbf{Appendix}\label{sec:appendix}}
\end{center}

\section{Qualitative Samples}\label{sec:appendix_qual}
We show a few randomly selected qualitative sample predictions from our model below. The model seems to perform well in most cases, particularly doing pretty well for named entities such as ``Gupta Empire'' and ``Emily Perkins''. Some of the failure cases include large contexts with references that need prior knowledge built into the system, an example of which would be for the question ``when was the first wonder woman comic released'' where the context refers to two dates for two events - (1) character first appeared in December 1941 (2) character's first cover was in January 1942. This is a particularly difficult question to answer as the model should understand that comic release date refers to when the character was first appeared on Sensation Comics cover.

\begin{tcolorbox}[breakable, enhanced]
    \textbf{Question:}the golden age of india took place during the rule of the \\
    \textbf{Context:} BPB The Gupta Empire was an ancient Indian empire , which existed at its zenith from approximately 319 to 485 CE and covered much of the Indian subcontinent . This period is called the Golden Age of India . The ruling dynasty of the empire was founded by Sri Gupta ; the most notable rulers of the dynasty were Chandragupta I , Samudragupta , and Chandragupta II . The 5th - century CE Sanskrit poet Kalidasa credits the Guptas with having conquered about twenty - one kingdoms , both in and outside India , including the kingdoms of Parasikas , the Hunas , the Kambojas , tribes located in the west and east Oxus valleys , the Kinnaras , Kiratas , and others . EEPE \\
    \textbf{Answer:} Gupta Empire \\
    \textbf{Prediction:} Gupta Empire
\end{tcolorbox}

\begin{tcolorbox}[breakable, enhanced]
    \textbf{Question:} Other than the motion picture and television industry, what other major industry is centered in Los Angeles? \\
    \textbf{Context:} The motion picture, television, and music industry is centered on the Los Angeles in southern California. Hollywood, a district within Los Angeles, is also a name associated with the motion picture industry. Headquartered in southern California are The Walt Disney Company (which also owns ABC), Sony Pictures, Universal, MGM, Paramount Pictures, 20th Century Fox, and Warner Brothers. Universal, Warner Brothers, and Sony also run major record companies as well.\\
    \textbf{Answer:} music\\
    \textbf{Prediction:} music industry
\end{tcolorbox}

\begin{tcolorbox}[breakable, enhanced]
    \textbf{Question:} Where was a lab for Tesla set up?\\
    \textbf{Context:} In late 1886 Tesla met Alfred S. Brown, a Western Union superintendent, and New York attorney Charles F. Peck. The two men were experienced in setting up companies and promoting inventions and patents for financial gain. Based on Tesla's patents and other ideas they agreed to back him financially and handle his patents. Together in April 1887 they formed the Tesla Electric Company with an agreement that profits from generated patents would go $1/3$ to Tesla, $1/3$ to Peck and Brown, and $1/3$ to fund development. They set up a laboratory for Tesla at 89 Liberty Street in Manhattan where he worked on improving and developing new types of electric motors, generators and other devices.\\
    \textbf{Answer:} Manhattan\\
    \textbf{Prediction:} 89 Liberty Street in Manhattan
\end{tcolorbox}

\begin{tcolorbox}[breakable, enhanced]
    \textbf{Question:} who plays unis in she 's the man\\
    \textbf{Context:} BLiB Emily Perkins as Eunice Bates , Olivia 's nerdy , eccentric friend and Duke 's lab partner EELiE\\
    \textbf{Answer:} Emily Perkins\\
    \textbf{Prediction:} Emily Perkins
\end{tcolorbox}

\begin{tcolorbox}[breakable, enhanced]
    \textbf{Question:} Why did Berlin Huguenots switch to German from French in their services?\\
    \textbf{Context:} In Berlin, the Huguenots created two new neighbourhoods: Dorotheenstadt and Friedrichstadt. By 1700, one-fifth of the city's population was French speaking. The Berlin Huguenots preserved the French language in their church services for nearly a century. They ultimately decided to switch to German in protest against the occupation of Prussia by Napoleon in 1806-07. Many of their descendents rose to positions of prominence. Several congregations were founded, such as those of Fredericia (Denmark), Berlin, Stockholm, Hamburg, Frankfurt, Helsinki, and Emden.\\
    \textbf{Answer:} in protest against the occupation of Prussia by Napoleon\\
    \textbf{Prediction:} protest against the occupation of Prussia by Napoleon in 1806-07
\end{tcolorbox}

\begin{tcolorbox}[breakable, enhanced]
    \textbf{Question:} How was scarcity managed in many countries?\\
    \textbf{Context:} Price controls exacerbated the crisis in the US. The system limited the price of "old oil" (that which had already been discovered) while allowing newly discovered oil to be sold at a higher price to encourage investment. Predictably, old oil was withdrawn from the market, creating greater scarcity. The rule also discouraged development of alternative energies. The rule had been intended to promote oil exploration. Scarcity was addressed by rationing (as in many countries). Motorists faced long lines at gas stations beginning in summer 1972 and increasing by summer 1973.\\
    \textbf{Answer:} rationing\\
    \textbf{Prediction:} rationing
\end{tcolorbox}

\begin{tcolorbox}[breakable, enhanced]
    \textbf{Question:} In which year did the newspaper define southern California?\\
    \textbf{Context:} In 1900, the Los Angeles Times defined southern California as including "the seven counties of Los Angeles, San Bernardino, Orange, Riverside, San Diego, Ventura and Santa Barbara." In 1999, the Times added a newer county—Imperial—to that list.\\
    \textbf{Answer:} 1900\\
    \textbf{Prediction:} In 1900
\end{tcolorbox}

\begin{tcolorbox}[breakable, enhanced]
    \textbf{Question:} when was the first wonder woman comic released\\
    \textbf{Context:} BPB Wonder Woman is a fictional superhero appearing in American comic books published by DC Comics . The character is a founding member of the Justice League , a goddess , and Ambassador - at - Large of the Amazon people . The character first appeared in All Star Comics \# 8 in December 1941 and first cover - dated on Sensation Comics \# 1 , January 1942 . In her homeland , the island nation of Themyscira , her official title is Princess Diana of Themyscira , Daughter of Hippolyta . When blending into the society outside of her homeland , she adopts her civilian identity Diana Prince . The character is also referred to by such epithets as the Amazing Amazon '' , the Spirit of Truth '' , Themyscira 's Champion '' , the God - Killer '' , and the `` Goddess of Love and War '' . EEPE\\
    \textbf{Answer:} January 1942\\
    \textbf{Prediction:} December 1941
\end{tcolorbox}

\end{document}